\documentclass[letterpaper, 10 pt, conference]{ieeeconf}  
\usepackage{graphicx}

\usepackage{float}
\usepackage{amsmath}
\usepackage{booktabs}

\usepackage{amsmath}
\usepackage{amssymb}
\IEEEoverridecommandlockouts 

\title{\LARGE \bf
Delay-Compensated Stiffness Estimation for Robot-Mediated Dyadic Interaction
}
\author{Mingtian Du$^{1}$, Suhas Raghavendra Kulkarni$^{1}$,  Bernardo Noronha$^{2}$, Domenico Campolo$^{1}$
\thanks{*This research was partially supported through the project "Empowering Remote Expertise: Advanced Haptic Technology for Enhanced Interaction" with funding support through NTUitive Gap Fund (POC) under Grant No. NGF-2024-16-019.}
\thanks{$^{1}$M. Du, S. R. Kulkarni, and D. Campolo are with Robotics Research Centre, School of Mechanical and Aerospace Engineering, Nanyang Technological University, Singapore 639798
        {\tt\small (corresponding author (D. Campolo): d.campolo@ntu.edu.sg).}}
\thanks{$^{2}$ B. Noronha is with Articares Pte. Ltd., 67 Ayer Rajah Crescent, \#07-11/12, Singapore 139950.}
}

\begin{document}

\maketitle
\begin{abstract}
Robot-mediated human-human (dyadic) interactions enable therapists to provide physical therapy remotely, yet an accurate perception of patient stiffness remains challenging due to network-induced haptic delays. Conventional stiffness estimation methods, which neglect delay, suffer from temporal misalignment between force and position signals, leading to significant estimation errors as delays increase. To address this, we propose a robust, delay-compensated stiffness estimation framework by deriving an algebraic estimator based on quasi-static equilibrium that explicitly accounts for temporally aligning the expert's input with the novice's response. A Normalised Weighted Least Squares (NWLS) implementation is then introduced to robustly filter dynamic bias resulting from the algebraic derivation. Experiments using commercial rehabilitation robots (H-MAN) as the platform demonstrate that the proposed method significantly outperforms the standard estimator, maintaining consistent tracking accuracy under multiple introduced delays. These findings offer a promising solution for achieving high-fidelity haptic perception in remote dyadic interaction, potentially facilitating reliable stiffness assessment in therapeutic settings across networks.
\end{abstract}

\section{Introduction}
Robot-mediated human-human (dyadic) interaction has emerged as a promising solution for facilitating remote haptic communication \cite{KucuktabakPons2021}, offering significant potential for telerehabilitation applications involving therapists and patients. In these architectures, a robotic platform mediates the haptic interaction between a skilled expert (therapist) and a less-skilled novice (patient) via a communication network. Previous studies have demonstrated that dyadic interaction outperforms individual task execution \cite{GaneshBurdet2014}, suggesting that skilled participants can effectively assist less skilled counterparts \cite{KagerCampolo2019}. However, in networked environments, the time delay is known to degrade the motor performance of dyads \cite{IvanovaBurdet2021,DuCampolo2024}, a phenomenon that could be attributed to controller instability induced by the delay \cite{DuCampolo2025}. A critical, yet under-investigated component of this interaction is the therapist's ability to accurately perceive the patient's physical state, specifically their muscle tone or stiffness. While the impact of time delay on motor performance is well-documented, the extent to which delay hinders accurate stiffness estimation has not been fully characterised.

Stiffness estimation is intrinsically linked to identifying the user's equilibrium position, a concept previously investigated for regulating force via spring-like muscle activation \cite{TakagiKoike2024}. Therefore, this estimation is often simplified as the correlation between the generated force and position displacement \cite{FangTsagarakis2018}. For the therapist, perceiving this mechanical stiffness constitutes a fundamental diagnostic modality essential for evaluating spasticity, rigidity, and rehabilitative progression \cite{Melendez-CalderonMussa-Ivaldi2013}. While such assessment traditionally relies on direct palpation in proximal settings, teleoperation architectures aim to preserve the haptic transparency during remote interaction. However, the fidelity of conventional stiffness estimation in telerehabilitation could be compromised by haptic delay induced by networks. 

Conventional ``Naive'' stiffness estimation techniques typically neglect communication delay, computing stiffness by correlating the instantaneous interaction force with delayed position feedback. As the time delay increases, the resulting temporal misalignment could induce significant estimation errors by potentially creating a temporal decoupling between the expert and the novice. Conversely, more sophisticated approaches, such as those utilising Communication Disturbance Observers to estimate network disturbances \cite{NatoriJezernik2010}, prioritise system stability and delay compensation over the accurate extraction of the novice's mechanical properties. To mitigate these limitations, this paper proposes a delay-compensated stiffness estimation framework for robot-mediated dyadic interaction. This estimator is derived from the quasi-static equilibrium of the delayed system and employs a force-based methodology by utilising the actual interaction force measured at the expert's interface.

To validate the proposed framework, experiments were conducted using a dyadic setup mediated by two commercial planar rehabilitation robots (Articares H-MAN). This platform served as a representative healthcare environment to systematically assess the impact of haptic delay on stiffness estimation. The analysis quantified the statistical degradation of the Naive estimator under increasing delays, demonstrating the detrimental effect of haptic delay on this reference model. Furthermore, the study evaluated the capacity of the proposed compensation method to maintain accuracy in the presence of significant delays. Consequently, the primary contribution of this work is the derivation of a force-based, delay-compensated estimator that explicitly compensates for round-trip haptic delays. Collectively, these findings provide a practical, computationally efficient tool to enhance the transparency of telerehabilitation systems, enabling therapists to reliably assess patient stiffness over networks.

\section{Stiffness Estimation Framework}
\begin{figure}
    \centering
    \vspace{0.4em} \includegraphics[width=0.8\linewidth]{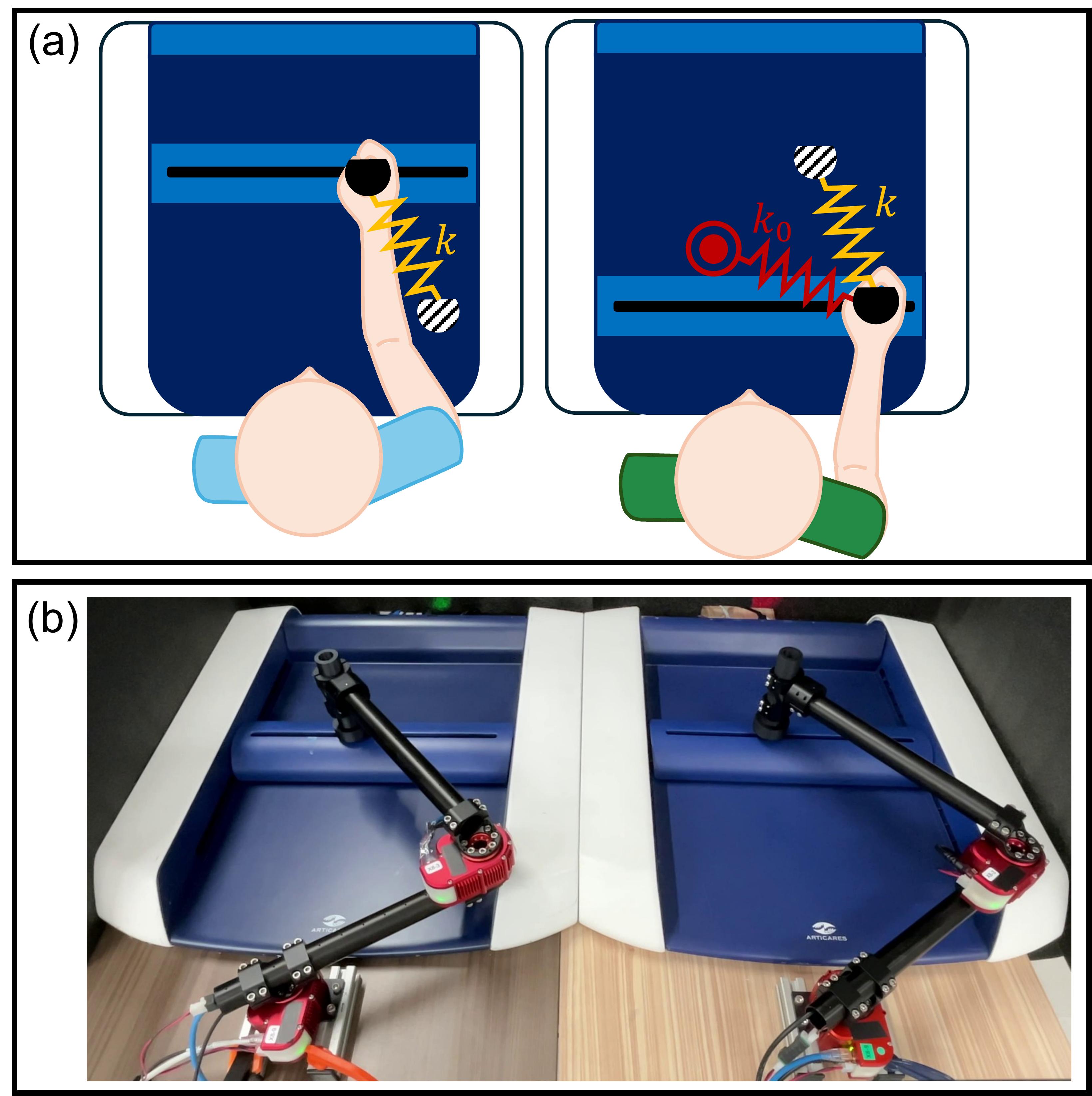}
    \caption{(a) Schematic representation of the stiffness estimation framework for robot-mediated dyadic interaction. The system couples an expert (therapist) and a novice (patient) via a virtual stiffness $k$, aiming to estimate the novice's intrinsic stiffness $k_0$ (representing muscle tone) in the presence of communication delay. (b) The physical experimental setup. Two planar H-MAN robots serve as the expert and novice interfaces, where human agents are replaced by HEBI actuator-based mechanisms to simulate programmable human joint impedance and ensure consistent ground truth generation.}
    \label{fig:setup}
\end{figure}

\begin{figure}
    \centering
    \vspace{0.4em}
    \includegraphics[width=0.8\linewidth]{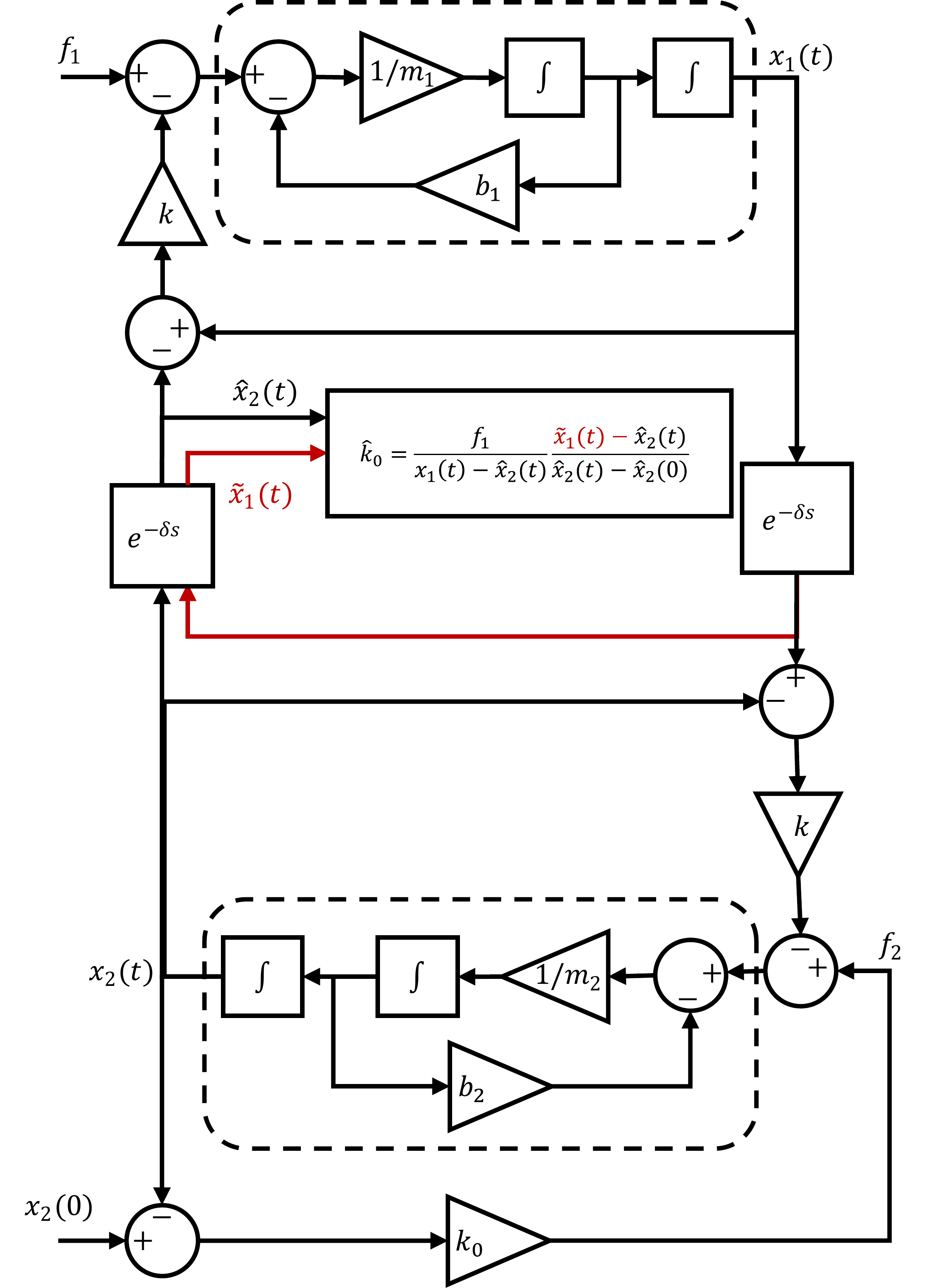}
    \caption{Block diagram of the haptic-delayed dyadic interaction system. The expert and novice interfaces are connected via a virtual coupling stiffness $k$ across a communication channel with haptic delay $\delta$. The diagram illustrates the signal causality, where the local expert force $f_1(t)$ and the delayed novice position $\hat{x}_2(t) = x_2(t-\delta)$ act as the observable variables for the stiffness estimator. Additionally, an observer on the expert side retrieves the round-trip delayed position $\tilde{x}_1(t) = x_1(t-2\delta)$ to enable the proposed delay-compensated estimation.}
    \label{fig:cd}
\end{figure}

The schematic representation of the proposed stiffness estimation framework is illustrated in Figure \ref{fig:setup}(a). In this model, the novice's stiffness is conceptualised as an elastic restoring force acting between the current end-effector position and the novice's intended target (equilibrium) position \cite{TakagiKoike2024}. Consequently, the robot-mediated dyadic interaction is modelled as a bilateral teleoperation system, as detailed in the control diagram in Figure \ref{fig:cd}. The dynamics governing the expert and novice robots are defined as:
\begin{equation}
    \begin{split}
        m_1\ddot{x}_1(t) + b_1\dot{x}_1(t) &= k(x_2(t-\delta) - x_1(t)) + f_1(t) \\
        m_2\ddot{x}_2(t) + b_2\dot{x}_2(t) &= k(x_1(t-\delta) - x_2(t)) + f_2(t)
    \end{split}
    \label{eq:cd}
\end{equation}
where the interaction force from the novice is modelled as a linear stiffness $f_2(t) = -k_0(x_2(t) - x_2(0))$. The primary objective of the framework in this study is to accurately estimate the novice's stiffness, denoted as $\hat{k}_0$, thereby providing the expert a reliable quantitative assessment. However, this estimation process is inherently challenged by communication delays, which introduce a temporal misalignment between the force and position signals.

Under quasi-static conditions, where inertial and damping effects are rendered negligible by the low velocity and acceleration of the system ($\dot{x} \approx 0$ and $\ddot{x} \approx 0$) \cite{MeriamBolton2020}, a force-based estimation strategy can be derived directly from the equilibrium of the system from Eq. \ref{eq:cd}. This approach utilises the interaction force $f_1(t)$ measured at the expert interface.
\subsection{The Naive Estimator}
Conventional stiffness estimation schemes often neglect propagation delays, assuming instantaneous signal transmission \cite{FangTsagarakis2018}. This ``Naive'' estimator correlates the instantaneous force measured at the expert side with the delayed position feedback received from the novice side. The resulting stiffness estimate, $\hat{k}_{0,\mathrm{naive}}$, is formulated as:
\begin{equation} \hat{k}_{0,\mathrm{naive}}(t) = \frac{f_1(t)}{\hat{x}_2(t) - \hat{x}_2(0)} \label{eq:naive} \end{equation}
where $\hat{x}_2(t)$ represents the novice's position as observed at the expert interface. In the presence of the haptic delay $\delta > 0$, this observed signal corresponds to the delayed state $\hat{x}_2(t) = x_2(t-\delta)$. For the initial condition, we assume the system rests at a static equilibrium prior to the start of the experiment ($t \le 0$), such that the initial observation $\hat{x}_2(0)$ corresponds identically to the novice's true initial position $x_2(0)$. However, during motion, the temporal misalignment between the instantaneous reaction force $f_1(t)$ and the delayed position $\hat{x}_2(t)$ could result in erroneous stiffness estimates, particularly during dynamic transients or direction reversals where the phase difference between the signals is most pronounced.

Direct implementation of Eq. \ref{eq:naive} risks numerical instability, particularly when the denominator approaches zero, potentially causing singularities. To mitigate this ill-conditioning, the estimator is reformulated as a linear regression problem that can be solved using the Least Squares (LS) method:
\begin{equation}
    \label{eq:naiveLS}
    \hat{k}_{0,\mathrm{naive}} = \underset{k}{\mathrm{argmin}}\sum_{t=0}^T\left[f_1(t) - k(\hat{x}_2(t) - \hat{x}_2(0))\right]^2
\end{equation} 
\subsection{Delay-Compensated Estimator: Ordinary Least Square}
To mitigate the estimation error induced by delay, an algebraic estimator based on the quasi-static equilibrium of the delayed system can be derived. The main challenge of this delay-compensation is that the real-time novice position $x_2(t)$ can not be obtained by the expert side, which can only observe the delayed position from the novice, i.e. $\hat{x}_2(t)$.

To address this challenge,  we reformulate the novice's quasi-static equilibrium by shifting the entire equation back by the delay $\delta$, relating the delayed response $\hat{x}_2(t)$ to the round-trip delayed expert command $x_1(t-2\delta)$. As shown in Figure \ref{fig:cd}, the controller can be augmented with an observer to retrieve the round-trip delayed position $\tilde{x}_1(t) = x_1(t-2\delta)$. Consequently, by relating the force exerted by the expert to the novice's stiffness through the system equilibrium, we obtain the delay-compensated estimator:
\begin{equation}
    \label{eq:cross_product}
    \hat{k}_{0,\mathrm{OLS}}(t) = \left[ \frac{f_1(t)}{x_1(t) - \hat{x}_2(t)} \right] \cdot \frac{\tilde{x}_1(t) - \hat{x}_2(t)}{\hat{x}_2(t) - x_2(0)}
\end{equation}
This formulation temporally aligns the input excitation (expert position) with the observed response (novice feedback). The implementation of Eq. \ref{eq:cross_product} can be rearranged as the linear regression expression solved by the Ordinary Least Squares (OLS) method.
\begin{equation}
\begin{split}
\label{eq:OLS}
\hat{k}_{0,\mathrm{OLS}}  &= \underset{k}{\mathrm{argmin}}\sum_{t=0}^T\left[Y(t) - k\Phi(t)\right]^2\\
    Y(t) &= f_1(t) \cdot \left[\tilde{x}_1(t) - \hat{x}_2(t)\right]\\
    \Phi(t) &= \left[x_1(t) - \hat{x}_2(t)\right] \cdot \left[\hat{x}_2(t) - \hat{x}_2(0)\right]
\end{split}
\end{equation}
Solving the Eq. \ref{eq:OLS} can minimise the squared error of $\sum (Y - \Phi \hat{k}_0)^2$. However, the transformation from Eq. \ref{eq:naiveLS} to Eq. \ref{eq:OLS} introduces a displacement-dependent weighting factor as in Figure \ref{fig:NWLS}. This implicit weighting biases the estimator towards high-displacement data points (Large $x_1 - \hat{x}_2$ or $\tilde{x}_1 - \hat{x}_2$), which typically correspond to high-acceleration phases. Since the quasi-static assumption neglects inertial dynamics ($m\ddot{x}$), these high-acceleration points may contain significant modelling errors. Therefore, a normalisation method is suggested to remove this bias.

\subsection{Implementation: Normalised Weighted Least Squares}
\begin{figure}
    \centering
    \vspace{0.4em}
    \includegraphics[width=0.8\linewidth]{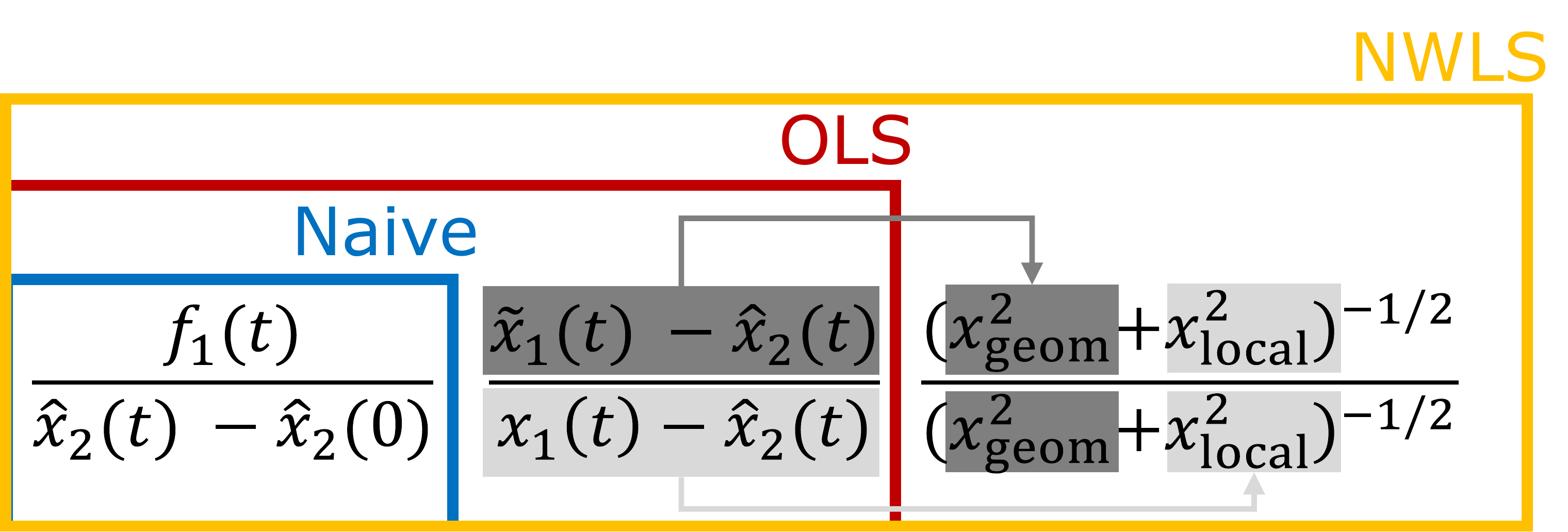}
    \caption{Ordinary Least Squares (OLS) add weights bias denoted by $x_\mathrm{geom} = \tilde{x}_1(t) - \hat{x}_2(t)$, and  $x_\mathrm{local} = x_1(t) - \hat{x}_2(t)$. When there is no delay, $x_\mathrm{geom} = x_\mathrm{local}$. Data points with a larger $x_\mathrm{geom}$ or $x_\mathrm{local}$ will be taken into regression with higher weights. Normalised Weighted Least Squares (NWLS) can remove this bias by placing a quadratic weight inversely proportional to the magnitude of the displacements on each data point, i.e. $(x^2_\mathrm{geom} + x^2_\mathrm{local})^{-1/2}$.}
    \label{fig:NWLS}
\end{figure}
To enforce uniform weights across the operational workspace, we employ a Normalised Weighted Least Squares (NWLS) scheme with a normalisation matrix $W$. The optimal estimate is given by:
\begin{equation}
    \hat{k}_{0, \mathrm{NWLS}} = \left( \Phi^\mathrm{T} W \Phi \right)^{-1} \Phi^\mathrm{T} W Y
\end{equation}
where $W$ is a diagonal weight matrix with entries $w_{ii}$, which are defined as the inverse of the deflection magnitude vector to counteract the implicit weighting introduced by the delay-compensation transformation:
\begin{equation}
w_{ii} = \left[\sqrt{(\tilde{x}_1(t) -\hat{x}_2(t))^2 + (x_1(t) - \hat{x}_2(t))^2} + \epsilon\right]^{-1}\\
\end{equation}
$\epsilon$ is a regularisation term to prevent division by zero. This normalisation introduces weights to counteract the bias inherent in Eq. \ref{eq:OLS}, as illustrated in Figure \ref{fig:NWLS}. This ensures that the estimator remains unbiased with respect to the interaction magnitude, preserving accuracy across both the static and dynamic phases of manipulation.

\section{Experimental Evaluation}
\subsection{System Description}
The performance of the stiffness estimator was evaluated through experimental validation. The setup consisted of two HEBI-actuated robotic interfaces coupled with two commercial planar rehabilitation robots (H-MAN) in Figure \ref{fig:setup}(b). This specific hardware configuration has been previously utilised to investigate the stability of dyadic interactions \cite{DuCampolo2025}. To generate a consistent excitation signal for estimation, the expert robot was commanded to track a sinusoidal trajectory along both the X and Y axes, governed by the following impedance control law:
\begin{equation}
    f_{1,i}^{\mathrm{cmd}} = k_c(p_{1,i}^* - p_{1,i}), \quad p_{1,i}^* = A\sin(\omega t)
\end{equation}
where $p_{1,i}^*$ represents the desired reference trajectory for the expert robot along X or Y axis ($i$ can be x or y). The control parameters were set to a stiffness of $k_c = 120\,$N/m, an amplitude of $A = 0.05\,$m, and an angular frequency of $\omega = 0.518\,\mathrm{rad/s}$, $\omega$ is selected to be significantly slower than a nominal human movement frequency exhibited in previous work \cite{TurlapatiCampolo2024}.
\begin{figure*}[thpb]
    \centering
    \vspace{0.4em} 
    \centering
    \includegraphics[width=7in]{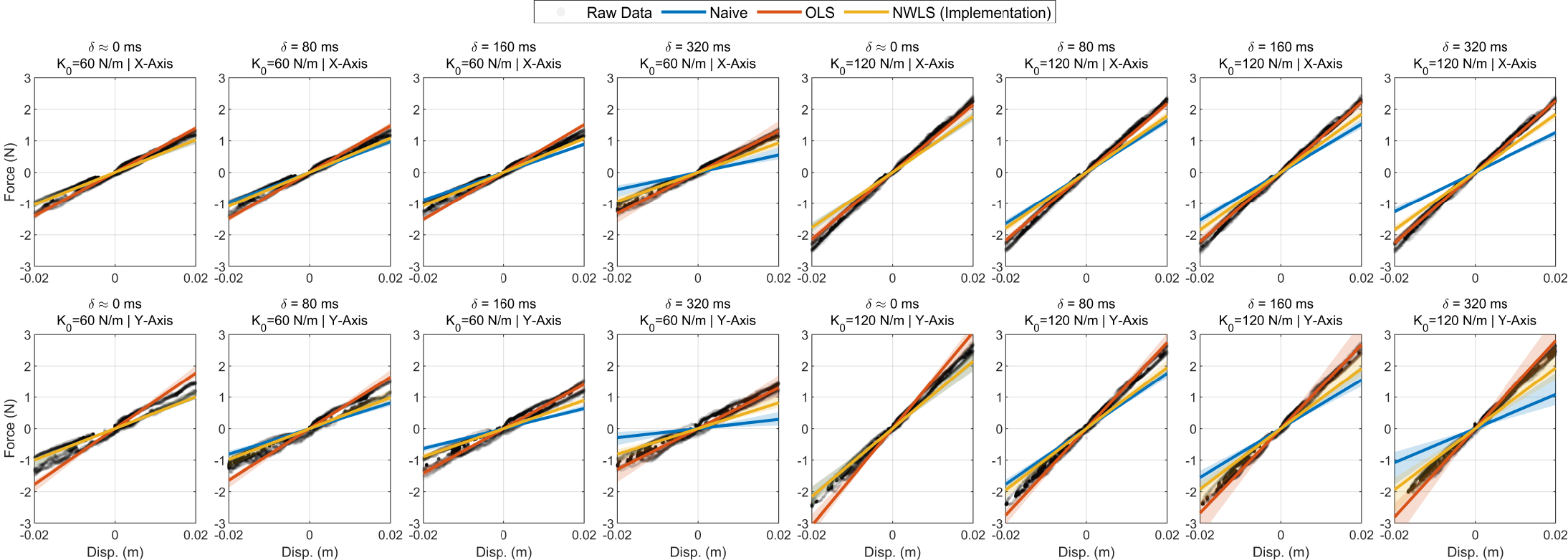}
    \caption{Comparative stiffness regression analysis across experimental conditions. Each subplot displays the force-displacement data (grey dots) collected from 10 independent trials for a specific combination of haptic delay ($\delta$), commanded novice stiffness ($\mathrm{K}_0$), and operational axis (X or Y-Axis). The coloured bands represent the range of stiffness estimates (minimum to maximum slope) generated by the Naive (blue), OLS (red), and Proposed NWLS (yellow) estimators, with the solid lines indicating the mean estimate. The alignment of these regression bands with the distribution of the raw force-displacement data provides a qualitative measure of estimation accuracy and robustness against dynamic transients. The displacement axes (X-Axis) limits are fixed to $\pm 0.02\,$m to maintain a consistent visual scale across stiffness conditions.}
    \label{fig:rawPlot}
\end{figure*}
\subsection{Experimental Protocol}
To rigorously evaluate the performance of the proposed delay-compensated estimator, we designed a comprehensive experimental protocol. The primary objective is to quantify the estimation accuracy of the novice's stiffness parameter ($\hat{k}_0$) under varying communication constraints. We employed a full factorial design that systematically varied the communication haptic delay and the mechanical stiffness of the novice robot. To verify the directional robustness of the estimator, this entire protocol was executed independently for both the X and Y operational axes of the H-MAN, a homogenous planar platform capable of decoupled actuation along orthogonal directions.

Throughout all experimental trials, the virtual coupling stiffness $k$ was maintained at a constant value of $200\,$N/m. The experiment investigated four distinct levels of haptic delays. These included a baseline condition with minimal system latency ($\approx$ 0 ms) to serve as an experimental control, and three introduced delays of $80\,$ms, $160\,$ms, and $320\,$ms. These values were chosen to simulate a range of network qualities, from standard regional connections to challenging intercontinental links where conventional Naive estimators could diverge. For each delay setting, the stiffness of the novice robot $k_0$ was set to two reference levels: $60\,$N/m and $120\,$N/m. These values were selected to mimic clinically relevant variations in human muscle tone, representing a hypotonic (relaxed) limb and a hypertonic (spastic) limb, respectively.

To ensure statistical robustness in the presence of sensor noise and mechanical friction variability, we conducted $10$ independent trials for each unique combination of delay, stiffness, and axis. This resulted in a total dataset of $160$ trials. 

\subsection{Data Analysis and Performance Metrics}
Because the physical stiffness rendered by the novice robot may deviate from the nominal commanded stiffness, the reference stiffness, denoted as $k_{0,\mathrm{ref}}$, was computed post-hoc for each experimental trial. This reference was derived by applying a standard LS fit to the recorded force feedback ($f_{2}$) and position feedback ($x_{2}$) data obtained directly from the novice robot's onboard sensors. This approach captures the true physical impedance encountered at the novice interface, encompassing both the active stiffness and any modelled passive dynamics:
\begin{equation}
    k_{0,\mathrm{ref}} = \underset{k}{\mathrm{argmin}} \sum_{t=0}^{T} \left[ f_2(t) - k (x_{2}(t) - x_2(0) \right]^2
\end{equation}
The proposed estimator operates on delayed and potentially distorted remote signals to yield the stiffness estimation $\hat{k}_{0,\mathrm{est}}$. To evaluate performance, we define the Absolute Percentage Error (APE) as the primary metric:
\begin{equation}
    E_{\mathrm{APE}} = \frac{|\hat{k}_{0,\mathrm{est}}  -k_{0,\mathrm{ref}} |}{|k_{0,\mathrm{ref}} |} \times 100\%
\end{equation}
where $\hat{k}_{0,\mathrm{est}}$ corresponds to the estimate from the Naive ($\hat{k}_{0,\mathrm{naive}}$), OLS ($\hat{k}_{0,\mathrm{OLS}}$), or NWLS ($\hat{k}_{0,\mathrm{NWLS}}$) method. This metric normalises the evaluation across varying stiffness levels ($60$ and $120$ N/m), delay intervals ($0$--$320$ ms), and operational axes (X and Y), providing a consistent measure of relative accuracy and directional robustness. Statistical significance was evaluated using the Wilcoxon rank sum test ($N=10$, $p < 0.05$).
\begin{figure}
    \centering
    \vspace{0.4em}
    \includegraphics[width=3.4in]{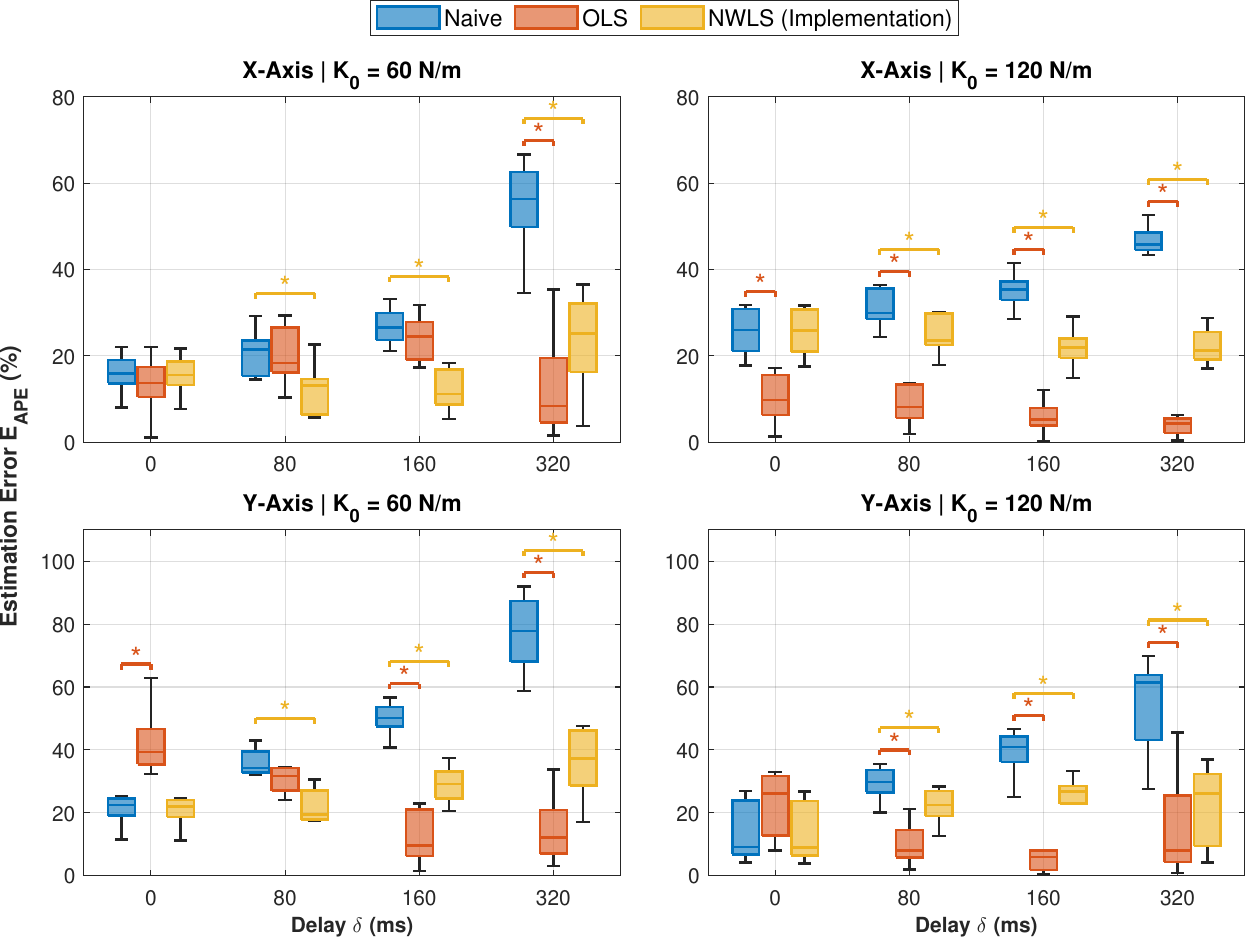}
    \caption{Statistical comparison of stiffness estimation accuracy. The boxplots illustrate the distribution of the Absolute Percentage Error ($E_{\mathrm{APE}}$) across 10 trials for each experimental condition, grouped by operational axis (X or Y-Axis) and novice stiffness ($\mathrm{K_0}$). The results compare the performance of the Naive (blue), OLS (red), and proposed NWLS (yellow) estimators across four delay conditions ($0$, $80$, $160$, and $320$ ms). An asterisk denotes a statistically significant difference ($p < 0.05$) between the OLS or NWLS estimator and the Naive method under identical conditions. Outliers are omitted for visual clarity.}
    \label{fig:boxPlot}
\end{figure}
\section{Experimental Results}
The performance of the proposed estimator was first evaluated qualitatively by examining the alignment of the estimated stiffness slopes with the raw interaction data, as shown in Figure \ref{fig:rawPlot}. This figure presents the force-displacement characteristics for all 16 experimental conditions, where the scatter of raw data points (grey) represents the physical ground truth of the interaction and the slope corresponds to the actual realised stiffness. Visual inspection suggests a divergence in the performance of the Naive estimator as the haptic delay increases. In conditions with larger delays (e.g., $\delta \ge 160$ ms), the Naive regression band (blue) appears to deviate below the trajectory of the raw data, implying an underestimation of the novice's stiffness.

In contrast, both the OLS (red) and the proposed NWLS (yellow) estimators appear to mitigate this misalignment. By shifting the regression bands back towards the density of the raw data, these methods seem to recover a slope that more closely follows the physical interaction. As illustrated in Figure \ref{fig:rawPlot}, the delay-compensated methods exhibit a visual consistency with the novice's force profile, whereas the Naive method shows a visible distortion, with the deviation appearing to increase with delay magnitude. To confirm these observations, a quantitative statistical analysis of the estimation error is presented in the following.

To quantify estimation accuracy, the APE was calculated for each trial relative to the measured ground truth stiffness. Figure \ref{fig:boxPlot} summarises the statistical distribution of these errors, highlighting distinct performance characteristics governed by the interaction between introduced delay and the estimation algorithm. The results demonstrate a clear degradation in the Naive estimator's performance as delay increases. Specifically, as the delay rises from $80\,$ms to $320\,$ms, the estimation error for the Naive method exhibits a substantial and monotonic increase across both the X and Y axes ($p<0.05$ for all pairwise comparisons between $80$, $160$, and $320\,$ms). Regarding the transition from baseline ($\approx0\,$ms) to $80\,$ms delay, the Y-Axis exhibited a statistically significant increase in estimation error (Naive, Y-Axis: $p<0.01$ for $0$ vs. $80\,$ms), whereas the X-Axis showed no significant difference (Naive, X-Axis: $p>0.05$ for $0$ vs. $80\,$ms).

The delay-compensated estimators (OLS and NWLS) demonstrate significant robustness to higher delay. When the haptic delay is set to the maximum ($320\,$ms), the OLS estimator yields significantly lower error compared to the Naive method ($320\,$ms: $p<0.05$ for OLS vs. Naive). However, at intermediate delays of $80$ and $160\,$ms, the OLS method does not consistently outperform the Naive approach (e.g., $160\,$ms, $60\,$N/m, X-Axis: $p>0.05$ for OLS vs. Naive). Notably, at baseline delay ($\approx 0\,$ms), the OLS estimator can perform even worse than the Naive method, yielding significantly larger errors in specific cases ($0\,$ms, $60\,$N/m, Y-Axis: $p<0.01$ for OLS vs. Naive). This degradation can be attributed to the implicit weighting of Eq. \ref{eq:OLS}, which disproportionately emphasises high-displacement data points, underscoring the necessity of the NWLS method. In contrast, the NWLS estimator consistently yields lower median errors compared to the Naive baseline across all conditions where time delay is present ($\ge 80\,$ms) ($80$-$320\,$ms: $p<0.05$ for NWLS vs. Naive). Furthermore, the NWLS method does not exhibit worse performance than the Naive method at the zero-delay baseline ($0\,$ms: $p>0.05$ for NWLS vs. Naive).

\section{Discussion}
The experimental analysis reveals that the Naive estimator, which neglects haptic delay, is significantly detrimented by the delay as small as $80\,$ms. A clear statistical trend indicates that the Naive estimate progressively degrades, exhibiting increasing underestimation as the delay rises, regardless of the stiffness level. These results underscore that a delay-compensation strategy is essential for accurate stiffness estimation in delayed haptic dyadic interactions. By aligning the locally measured force with the geometrically corrected position difference, the proposed framework demonstrates the capacity to mitigate the negative impact of significant delays.

While the OLS estimator outperforms the Naive method under significant delay (i.e., $320\,$ms), it exhibits performance degradation in the zero-delay baseline condition, specifically along the Y-Axis with the lower stiffness setting ($\delta \approx 0$, $\mathrm{K_0 = 60}$). This result could be attributed to the implicit weighting inherent in the derived delay-compensated estimator in Eq. \ref{eq:OLS}, which prioritises high-displacement regions corresponding to the turnaround points of the sinusoidal motion. At these points, velocity is near zero, making static friction dominant, while maximal acceleration introduces significant inertial forces. In the low-stiffness condition, the restorative spring force is relatively weak compared to these friction and inertial forces. Therefore, the OLS estimator can lead to an overestimation by focusing on these high-deflection regions. 

In contrast, the proposed NWLS method applies a normalisation weight that counteracts this bias by distributing importance evenly across the trajectory. This explains why NWLS maintains accuracy comparable to the Naive baseline at zero delay while effectively compensating for delays ($80$, $160$, and $320\,$ms) relative to the Naive method. Notably, NWLS does not consistently outperform OLS in all conditions (e.g., X-Axis, $\mathrm{K_0 = 120}$). This observation suggests a trade-off governed by the signal-to-noise ratio (SNR). Although the turnaround points contain inertial artefacts (inertial force $ma$) that violate the quasi-static assumption, they also correspond to the highest interaction forces, which maximise the SNR relative to sensor noise. The OLS estimator's implicit focus on these high-force regions, particularly when the stiffness signal is strong (i.e., $\mathrm{K_0 = 120}$), can therefore yield lower variance. However, this relies on a specific dynamic condition; the NWLS method, by enforcing uniform weighting, provides superior robustness across varying delays and stiffness levels, ensuring reliable estimation even when the specific bias of OLS is detrimental.

Although the X and Y axes of the H-MAN are programmed independently, they do not exhibit identical performance characteristics. The Y-Axis demonstrated higher APE ranges and instances where the OLS estimator performed worse than the Naive method, behaviours not observed in the X-Axis. These discrepancies can be attributed to the higher effective inertia along the Y-Axis, a known mechanical feature of the H-MAN platform \cite{CampoloMasia2014}. Furthermore, the high APE measured for the Naive estimator at baseline delay ($\approx 0\,$ms) is primarily driven by hardware limitations inherent to the actuators, sensors, and transmission mechanics. Despite these hardware-induced offsets, the estimator successfully tracked the step change from hypotonic ($60\,$N/m) to hypertonic ($120\,$N/m) conditions across all delay settings. This suggests that the system retains the sensitivity required to distinguish changes in novice stiffness, indicating potential applicability for detecting clinically relevant variations such as patient muscle tone.

The experimental validation in this study relied on constant haptic delays to isolate the fundamental effects of latency. Real-world networks, however, introduce stochastic characteristics such as packet jitter and loss. Future work will extend this framework to accommodate variable delays. Furthermore, while the NWLS estimator operates as a batch process, it can be naturally extended to a Recursive Least Squares formulation to enable real-time parameter tracking, which is essential for supervised remote manipulation scenarios. Finally, this study utilised a robot-on-robot configuration to ensure ground truth consistency. The next phase of research will involve trials with human subjects to evaluate the estimator's performance when interacting with the complex, non-linear, and time-varying impedance of biological limbs.

\section{Conclusion}

This work presents a delay-compensated stiffness estimation framework designed to enhance transparency in robot-mediated dyadic interactions. By deriving a force-based algebraic estimator grounded in quasi-static equilibrium that explicitly compensates for round-trip haptic delay, we resolve the temporal misalignment that compromises standard Naive estimation. Experimental validation on a rehabilitation robotic platform confirmed that the performance of the Naive estimator degrades significantly as the delay increases. In contrast, the proposed Normalised Weighted Least Squares implementation demonstrated robust performance, significantly reducing estimation error and maintaining consistent stiffness tracking under delays of up to $320\,$ms. These findings establish a viable pathway for high-fidelity haptic perception in remote interactions, potentially enabling therapists to perform reliable stiffness assessments over high-latency networks.

\bibliographystyle{IEEEtran}
\bibliography{reference}
\end{document}